\documentclass{article} 
\usepackage{iclr2024_conference,times}

\usepackage{amsmath,amsfonts,bm}

\def\eqref#1{equation~\ref{#1}}

\def\1{\bm{1}}

\def\rvc{{\mathbf{c}}}

\def\rvz{{\mathbf{z}}}

\DeclareMathAlphabet{\mathsfit}{\encodingdefault}{\sfdefault}{m}{sl}
\SetMathAlphabet{\mathsfit}{bold}{\encodingdefault}{\sfdefault}{bx}{n}

\usepackage[T1]{fontenc}
\usepackage[colorlinks=true, urlcolor=blue, citecolor=black, linkcolor=magenta]{hyperref}

\usepackage{graphicx}
\usepackage{subcaption}
\usepackage{wrapfig}
\usepackage{booktabs}
\usepackage{multirow}
\usepackage{lipsum}
\usepackage[font=small]{caption}

\definecolor{todooutline}{RGB}{0, 102, 204}
\definecolor{todobg}{RGB}{222, 235, 247}

\title{Learning to Act without Actions}

\author{Dominik Schmidt\thanks{Correspondence to 
\href{mailto:dominik.schmidt.22@ucl.ac.uk}{dominik.schmidt.22@ucl.ac.uk}
.}\\
Weco AI\\
\And
Minqi Jiang \\
FAIR at Meta AI \\
}

\newcommand{\method}[0]{LAPO}
\newcommand{\methodfull}[0]{Latent Action Policies} 

\iclrfinalcopy 
\begin{document}

\maketitle

\begin{abstract}

Pre-training large models on vast amounts of web data has proven to be an effective approach for obtaining powerful, general models in domains such as language and vision. However, this paradigm has not yet taken hold in reinforcement learning. This is because videos, the most abundant form of embodied behavioral data on the web, lack the action labels required by existing methods for imitating behavior from demonstrations. We introduce \textbf{\mbox{\methodfull{}}}~(\method{}), a method for recovering latent action information---and thereby latent-action policies, world models, and inverse dynamics models---purely from videos. \method{} is the first method able to recover the structure of the true action space just from observed dynamics, even in challenging procedurally-generated environments. \method{} enables training latent-action policies that can be rapidly fine-tuned into expert-level policies, either offline using a small action-labeled dataset, or online with rewards. \method{} takes a first step towards pre-training powerful, generalist policies and world models on the vast amounts of videos readily available on the web. Our code is available here: 
\url{https://github.com/schmidtdominik/LAPO}.

\end{abstract}

\section{Introduction}

Training on web-scale data has shown to be an effective approach for obtaining powerful models with broad capabilities in domains including language and vision~\citep{gpt2, dino}. Much recent work has thus sought to apply the same paradigm in deep reinforcement learning~\citep[RL,][]{sutton2018reinforcement}, in order to learn generalist policies from massive amounts of web data~\citep{vpt, gato}. However, common methods for learning policies from offline demonstrations, such as \emph{imitation learning}~\citep{pomerleau1988alvinn,ross2010efficient} and \emph{offline RL}~\citep{kumar2020conservative,levine2020offline}, typically require action or reward labels, which are missing from purely observational data, such as videos found on the web. 

In this work we introduce \textbf{\methodfull{}}~(\method{}), a method for recovering latent-action information from videos. In addition to inferring the structure of the underlying action space purely from observed dynamics, \method{} produces latent-action versions of useful RL models including forward-dynamics models (world models), inverse-dynamics models, and importantly, policies.

\method{} is founded on the key insights that (1) some notion of a \emph{latent action} that explains an environment transition can be inferred from observations alone, and (2) given such inferred latent actions per transition, a \emph{latent-action policy} can be obtained using standard imitation learning methods. Our experiments provide strong evidence that such latent policies accurately capture the observed expert's behavior, by showing that they can be efficiently fine-tuned into expert-level policies in the true action space. This results makes \method{} a significant step toward pre-training general, rapidly-adaptable policies on massive action-free demonstration datasets, such as the vast quantities of videos available on the web.

Crucially, \method{} learns to infer latent actions, and consequently, obtain latent action policies in a fully unsupervised manner. \method{} is similar to prior work~\citep{bco, rl_with_videos_idmtransfer_schmeck, vpt, ssorl} in that we first train
an \emph{inverse dynamics model} (IDM) that predicts the action taken between two consecutive observations, and then use this IDM to add action labels to a large action-free dataset. However, unlike these prior works, which rely on some amount of ground-truth action-labelled data for training the IDM, \method{} does not make use of any labels and infers latent action information purely from observed environment dynamics.
To do so, the IDM in \method{} learns to predict \emph{latent actions} rather than true actions. These latent actions take the form of a learned representation that explains an observed transition. To learn this representation, \method{} makes use of a simple unsupervised objective that seeks to establish predictive consistency between the IDM and a separate \emph{forward dynamics model} (FDM). The FDM is trained on transitions consisting of two consecutive observations $(o_t, o_{t+1})$ to predict the future observation $o_{t+1}$, given the past observation $o_t$ and a latent action. This latent action is generated by the IDM given both past and future observations.
Thus, unlike the FDM which sees only the past, the IDM has access to both the past and future and learns to pass useful information about the future to the FDM through the latent action. By making the latent action an information bottleneck~\citep{tishby2000information}, we prevent the IDM from simply forwarding the entire future observation to the FDM, thereby forcing it to learn a highly compressed encoding of state transitions. We refer to this encoding as a latent action, as it captures the observable effects of the agent's true actions.

\begin{figure}[t!]
\begin{center}
\vspace{-12mm}
\includegraphics[width=1.0\textwidth]{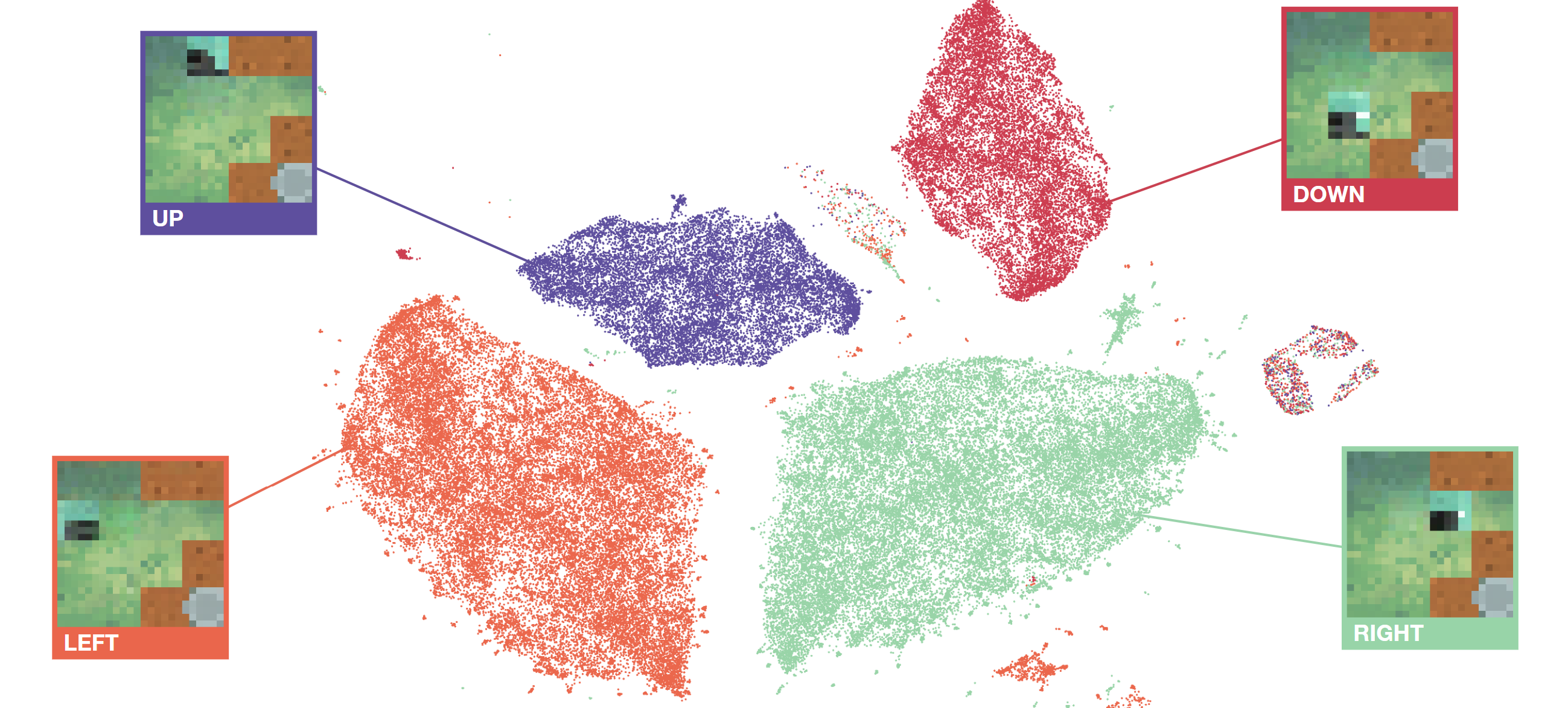}
\end{center}
\caption{\small{UMAP projection of the learned latent action space for \texttt{Miner} alongside illustrative next-state predictions generated by the FDM for each cluster of latent actions. Each point represents a continuous latent action generated by the IDM for a transition in the video dataset. Each point is color-coded by the true action taken by the agent at that transition. For clarity, \texttt{NOOP} actions are omitted. The structure of the latent action space is highly interpretable and closely corresponds to the true action space, even though no ground-truth action labels were used during training.
}}
\label{fig:intro_umap}
\vspace{-5mm}
\end{figure}

In our experiments in Section~\ref{sec:results}, we train a latent IDM via \method{} on large expert-level action-free offline datasets for each of the 16 games of the Procgen Benchmark~\citep{cobbe2019quantifying, procgen}. We observe that the structure of the learned latent action spaces are highly interpretable, often exhibiting clusters that closely align with the true action space~(see Figure \ref{fig:intro_umap}). This is a remarkable result, as our approach uses zero ground-truth action labels and instead recovers all action information purely from observed dynamics. To go beyond merely recovering action information, we then show how we can leverage the learned latent representation to quickly obtain powerful expert-level policies in downstream tasks: We first use the IDM to assign latent action labels to each transition in the same video dataset on which it was trained. We then perform behavior cloning on the resulting action-observation dataset to learn a latent-action policy, which imitates the latent actions predicted by the IDM. Finally, to deploy the policy in the online environment, we seek to turn it from a latent action policy into a policy in the true action space. We show that if a tiny, action-labeled dataset is available, a latent $\rightarrow$ true action decoder can be learned in a supervised manner and used to decode the outputs of the latent policy. This approach is extremely data-efficient, requiring only $\sim$200 labeled transitions to exceed the performance of 4M steps of regular RL via PPO. Alternatively, if an online environment is available, we show that simply fine-tuning the last layers of the latent policy with a general-purpose RL algorithm allows the policy to rapidly adapt to the true action space and recover (or at times exceed) expert performance. These results indicate that latent policies produced by \method{} capture meaningful behavior in the latent action space.

While we take a first step in demonstrating the potential of these latent representations for obtaining powerful policies in Section~\ref{sec:results}, the \textbf{fundamental contribution of this work} is simply showing that comprehensive action information can be recovered from pure video through unsupervised objectives, such as the one proposed herein. This result opens the door to future work that employs representation learning methods like \method{} to train powerful, generalist policies and world models on massive video datasets from the web.

\section{Related work}
\label{sec:related_work}

Most closely related to our approach, ILPO \citep{ilpo} aims to infer latent actions via a dynamics modelling objective. However, unlike \method{}, ILPO learns a latent policy rather than an IDM jointly with the FDM, and uses a discrete rather than a continuous latent action space. For each FDM update, ILPO identifies the discrete latent action that minimizes the next-state prediction error, and only optimizes the FDM for that specific action. In practice, this objective can lead to mode collapse, where a feedback loop causes only a small number of discrete latents to ever be selected \citep{ilpo_followup}. By using an IDM rather than a discrete latent policy, \method{} avoids enumerating all hypothetically possible transitions and instead learns the latent action corresponding to the actual observed transition. To train the policy, ILPO minimizes the difference between the true next state and the expectation of the next state under the policy. This loss is ill-conditioned, as even when it is minimized, the next-state predictions for individual latent actions do not have to align with the environment. Moreover, \method{}'s use of continuous rather than finite discrete latents may allow better modeling in complex and partially-observable environments, as any information useful for prediction can be captured within the latent representation. We provide evidence for this in Appendix~\ref{apdx:ilpo_repro}, where we show that ILPO breaks down when applied to more than a single Procgen level at a time. Finally, ILPO's use of discrete latents makes its computational complexity linear rather than constant in the latent dimensionality, due to the required enumeration of all actions.

In the model-based setting, FICC~\citep{latent_actions_for_mbrl} pre-trains a world model from observation-only demonstrations. FICC thus seeks to distill dynamics from demonstrations, rather than behaviors, as done by \method{}. Moreover, while the cyclic consistency loss used by FICC closely resembles the \method{} objective, the adaptation phase of FICC must operate in reverse, mapping true actions to latent actions in order to finetune the world model. Unlike the approach taken by \method{}, this adaptation strategy cannot make direct use of online, ground-truth action labels, and requires continued training and application of a world model. In contrast, \method{} directly produces latent action policies imitating the expert behavior, which can be rapidly adapted online to expert-level performance. Outside of RL, Playable Video Generation \citep[PVG,][]{pvg} similarly focuses on world model learning with a similar objective for the purpose of controllable video generation. Like ILPO, PVG uses a more limiting, small set of discrete latent actions. 

A related set of semi-supervised approaches first train an IDM using a smaller dataset containing ground-truth action labels, and then use this IDM to label a larger action-free dataset, which can subsequently be used for behavior cloning. VPT~\citep{vpt} and ACO~\citep{DBLP:conf/eccv/ZhangPZ22} follow this approach, training an IDM on action-labeled data which is then used to generate pseudo labels for training policies on unlabeled footage gathered from the web.
RLV~\citep{rl_with_videos_idmtransfer_schmeck} uses an observation-only dataset within an online RL loop, where action and reward labels are provided by an IDM trained on action-labelled data and a hand-crafted reward function respectively. Similarly, BCO~\citep{bco} trains an IDM through environment interaction, then uses the IDM to label action-free demonstrations for behavior cloning. However, relying on interactions for labels can be inefficient, as finding useful labels online may itself be a difficult RL exploration problem. SS-ORL~\citep{ssorl} is similar to VPT but performs offline RL instead of imitation learning on the IDM-labelled data and requires reward-labels. Unlike these methods, \method{} avoids the need for significant amounts of action-labeled data to train an IDM, by directly inferring latent actions and latent policies which can easily be decoded into true actions.

\method{} differs from previous methods for \emph{imitation learning from observation}~\citep[IfO;][]{ifo_survey, torabi2018generative, yang2019imitation}, which typically require the imitating policy to have access to the true action space when training on action-free demonstrations. Crucially, unlike these prior approaches, \method{} does not require access to the ground-truth action space to learn from action-free demonstrations. Other methods leverage observation-only demonstrations for purposes other than learning useful behaviors. Intention-Conditioned Value Functions~\citep[ICVF;][]{icvf} uses a temporal difference learning objective based on observation-only demonstrations to learn state representations useful for downstream RL tasks. \citet{hard_expl_youtube} propose to use expert demonstrations during online RL to guide the agent along the expert's trajectories. For this, they propose two auxiliary classification tasks for learning state representations based on observation-only demonstrations.

\raggedbottom
\newpage

\section{Background}
\label{sec:bg}

\subsection{Reinforcement Learning}
Throughout this paper, we consider RL under partial observability, using the framework of the partially-observable Markov decision process ~\citep[POMDP,][]{aastrom1965optimal,kaelbling1998planning}. A POMDP consists of a tuple $(\mathcal{S}, \mathcal{A}, \mathcal{T}, \mathcal{R}, \mathcal{O}, \Omega, \gamma)$, where $\mathcal{S}$ is the state space, $\mathcal{A}$ is the action space, $\mathcal{O}$ is the observation space, and $\gamma$ is the discount factor. At each timestep $t$, the RL agent receives an observation $o_t$ derived from the state $s_t$, according to the observation function $\Omega:\mathcal{S}\mapsto\mathcal{O}$, and takes an action according to its policy $\pi(a_t|o_t)$, in order to maximize its expected discounted return, $\sum_{k=t}^{\infty} \gamma^{k-t} r_k$. The environment then transitions to the next state $s_{t+1}$ according to the transition function, $\mathcal{T}: \mathcal{S}\times\mathcal{A} \mapsto \mathcal{S}$, and agent receives a reward $r_t$ based on the reward function $\mathcal{R}:\mathcal{S}\times\mathcal{A}\times\mathcal{S}\mapsto\mathbb{R}$. This work considers first learning a policy $\pi$ from offline demonstrations, followed by further fine-tuning the policy online as the agent interacts with its environment.

\subsection{Learning from Observations}

Often, we have access to recordings of a performant policy, e.g. a human expert, performing a task of interest. When the dataset includes the action labels for transitions, a supervised learning approach called \emph{behavior cloning}~\citep[BC,][]{pomerleau1988alvinn} can be used to train an RL policy to directly imitate the expert. Consider a dataset $D$ of such expert trajectories within the environment, where each trajectory $\tau$ consists of a list of all transition tuples $(o_0, a_0, o_1), \ldots, (o_{|\tau|-1}, a_{|\tau|-1}, o_{|\tau|})$ in a complete episode within the environment. BC then trains a policy $\pi_{\text{BC}}$ to imitate the expert by minimizing the cross-entropy loss between the policy's action distribution and the expert's action $a^*$ for each observation in $D$: $\mathcal{L}_{\text{BC}} = -\frac{1}{|D|}\sum_{\tau \in D}\sum_{t=0}^{|\tau|}\log(\pi(a_t^*|o_t))$. 

Unfortunately, most demonstration data in the wild, e.g. videos, do not contain action labels. In this case, the demonstration data simply consists of a continuous stream of observations taking the form $(o_0, o_1, \ldots, o_{|\tau|})$. This setting, sometimes called \emph{imitation learning from observations}~\citep[IfO,][]{ifo_survey}, poses a more difficult (and practical) challenge. IfO methods often seek to learn a model that predicts the missing action labels, typically trained on a separate dataset with ground-truth actions. Once labeled this way, the previously action-free dataset can then be used for BC. In this work, we 
likewise convert the IfO problem into the BC problem. However, instead of relying on access to a dataset with ground-truth action labels, our method directly infers \emph{latent actions} $z_t$ that explain each observed transition $(o_t, o_{t+1})$, with which we train \emph{latent action policies}, $\tilde{\pi}(z_t|o_t)$.

\subsection{Dynamics Models}
\method{} employs two kinds of dynamics models: The first is the \emph{inverse dynamics model}, $p_\text{IDM}(a_t | o_t, o_{t+1})$, which predicts which action $a_t$ was taken by the agent between consecutive observations $o_t$ and $o_{t+1}$. An IDM can be used to label a sequence of observations with the corresponding sequence of actions. The second is the \emph{forward dynamics model}, $p_\text{FDM}(o_{t+1} | o_t, a_t)$, which predicts the next observation $o_{t+1}$ given the previous observation $o_t$ and the action $a_t$ taken by the agent after observing $o_t$. The FDM can be used as an approximation of the environment's transition function. In this work, the IDM and FDM are deterministic and implemented as deep neural networks. Unlike a standard IDM, the IDM used by \method{} predicts continuous latent actions, $z_t \in \mathcal{Z}$, where $\mathcal{Z} = \mathbb{R}^n$.

\subsection{Vector-quantization}
Vector-quantization~\citep[VQ]{vq} is a method for learning discrete features by quantizing an underlying continuous representation. This has been shown particularly useful in deep learning, where VQ enables learning discrete representations while allowing gradients to flow through the quantization step. VQ has been used effectively in many domains including vision \citep{vqvae, vq_for_img, straightening_out_vq}, audio \citep{vq_audio_jukebox, vq_soundstream}, and model-based RL~\citep{iris}. To quantize a continuous vector $\rvz$, VQ maintains a codebook $\{\rvc_1, \rvc_2, \ldots, \rvc_m\}$ and maps $\rvz$ to the closest codebook vector $\rvc_i$. 
The straight-through gradient estimator is used to pass gradients through the quantization step \citep{straight_through_est}.

\section{\methodfull{}}
\label{sec:method}
\subsection{Learning a Latent Action Representation}
\begin{wrapfigure}{r}{0.32\textwidth}
    \vspace{-5mm}
    \centering
    \includegraphics[width=0.32\textwidth]{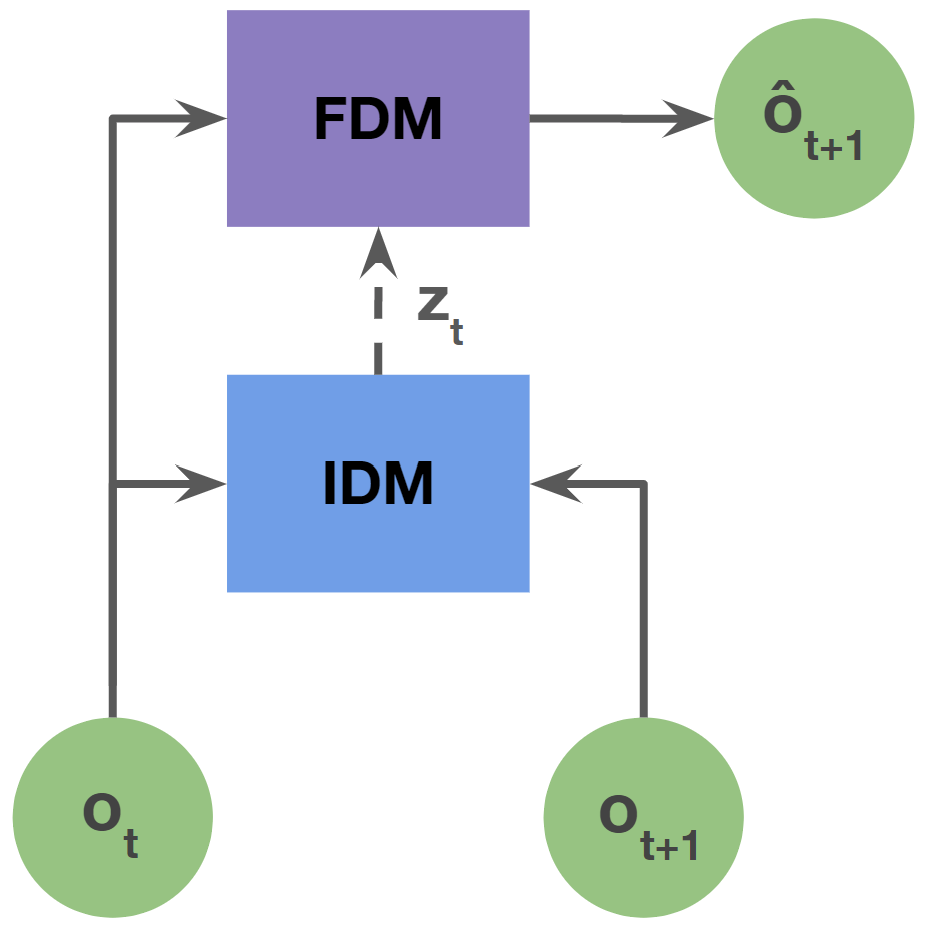}
    \caption{\small{\method{} architecture. Both IDM and FDM observe $o_t$, but only the IDM observes $o_{t+1}$. To enable accurate predictions of $o_{t+1}$, the IDM must pass useful transition information through the quantized information bottleneck $z_t$ to the FDM.}}
    \label{fig:lfo_diag}
    \vspace{5mm}
\end{wrapfigure}
We now describe our approach for learning a latent IDM via a forward dynamics modelling objective (see Fig. \ref{fig:lfo_diag}). First, to learn about the action at time $t$, we sample a sequence of observations $o_{t-k}, \ldots, o_{t}, o_{t+1}$ from our observation-only dataset. Here, $o_{t}$ and $o_{t+1}$ are the observations before and after the action of interest is taken and observations $o_{t-k}, \ldots, o_{t-1}$ are additional context controlled by hyperparameter $k \ge 0$. The IDM then predicts the latent action $z_t$ based on the full sequence of observations. $$z_t \sim p_\text{IDM}(\cdot | o_{t-k}, \ldots, o_{t}, o_{t+1})$$ 
Next, the FDM predicts the post-transition observation $o_{t+1}$ based only on the past observations $o_{t-k}, \ldots, o_{t}$ and the latent action $z_t$. $$\hat o_{t+1} \sim p_\text{FDM}(\cdot | o_{t-k}, \ldots, o_{t}, z_t)$$ Both models are trained jointly via gradient descent to minimize the next state prediction error $||\hat o_{t+1} - o_{t+1}||^2$.

This method, as described so far, would likely result in the IDM learning to simply copy $o_{t+1}$ into $z_t$, and the FDM learning to forward $z_t$ as is. To remedy this, we make the latent action an information bottleneck. This forces the IDM to compress any information to be passed to the FDM. Since both the FDM and IDM have access to past observations but only the IDM observes the post-transition observation $o_{t+1}$, the IDM is learns to encode only the \emph{difference} between $o_{t+1}$ and $o_t$ into $z_t$, rather than full information about $o_{t+1}$. Naturally, one highly efficient encoding of the differences of two consecutive observations, at least in deterministic environments, is simply the agent's true action. Our hypothesis is thus, that forcing the IDM to heavily compress transition information as described, may allow us to learn a latent action representation with a structure closely corresponding to the true action space.

As both the IDM and FDM have access to past observations, the learned latent actions may become conditional on these observations. Intuitively, this means that some latent action $z$ could correspond to different true actions when performed in different states. While this is not necessarily an issue, biasing the method toward simpler representations is likely preferrable~\citep{solmonoff1964formal,schmidhuber1997discovering,hutter2003existence}. Consequently, we apply vector quantization (VQ) to each latent action before passing it to the FDM, thus forcing the IDM to reuse the limited number of discrete latents across different parts of the state-space, leading to disentangled representations.

\subsection{Behavior Cloning a Latent Action Policy}
Using the trained latent IDM, we now seek to obtain a policy. For this, we initialize a latent policy $\pi\colon \mathcal{O} \rightarrow \mathcal{Z}$ and perform behavior cloning on the same observation-only dataset that the IDM was trained on, with the required action labels generated by the IDM. This is done via gradient descent with respect to policy parameters on the loss $||\pi(o_t) - z_t||^2$ where $z_t \sim p_\text{IDM}(\cdot | o_{t-k}, \ldots, o_{t}, o_{t+1})$.

\subsection{Decoding Latent Actions}
The policy $\pi$, obtained via BC, produces actions in the latent action space. We now consider different ways to adapt $\pi$ to produce outputs in the true action space, depending on the kinds of data available.

\paragraph{Small action-labeled dataset.} If a small dataset of true-action-labeled transitions is available, we can train an action decoder, $d\colon \mathcal{Z} \rightarrow \mathcal{A}$, to map the IDM-predicted latent action for each transition to the known ground-truth action. The trained decoder can then be composed with the frozen latent policy and the resulting decoded latent policy, $d \circ \pi\colon \mathcal{O} \rightarrow \mathcal{A}$, can be deployed online.

\paragraph{Online environment.} By interacting with the online environment we get both action-labeled transitions, which can be used to train a decoder in a supervised manner as described above, and a reward signal, which can be used for online RL. By doing both simultaneously, we can quickly bootstrap a crude, initial decoder via supervised learning, and then fine-tune the entire composed policy $d \circ \pi$ by directly optimizing for the environment's reward function via RL. Through RL, the performance of the policy can be potentially be improved beyond that of the data-generating policy. In particular, when the data was generated by a mixture of policies of different skill levels or pursuing different goals (as is the case with data from the web), this step can extract the data-generating policies more closely aligned with the reward function used for fine-tuning.

\section{Experimental Setting}
\label{sec:results_disc}

Our experiments center on the Procgen Benchmark~\citep{procgen}, as it features a wide variety of tasks that present different challenges for our method. Compared to other classic benchmarks such as Atari~\citep{bellemare2013arcade}, the procedural generation underlying Procgen introduces a difficult generalization problem and results in greater visual complexity which makes dynamics modeling at pixel-level challenging. Moreover, several Procgen environments feature partial observability, which along with stochasticity, presents issues for methods attempting to infer actions purely from observation by making it ambiguous which parts of an environment transition are due to the agent and which are due to stochastic effects or unobserved information. Our observation-only dataset consists of approximately 8M frames sampled from an expert policy that was trained with PPO for 50M frames. The use of this synthetic dataset, rather than video data from the web, allows us to better evaluate our method, as we can directly access the expert policy's true actions, as well as metadata such as episodic returns for evaluation purposes.

We use the IMPALA-CNN~\citep{impala} to implement both our policy and IDM with a 4x channel multiplier as used by \citet{procgen}, and U-Net~\citep{ronneberger2015unet} based on a ResNet backbone~\citep{resnet} with approximately 8M parameters for the FDM. The latent action decoder is a fully-connected network with hidden sizes $(128, 128)$. We use an EMA-based update~\citep{polyak1992acceleration} for the vector quantization embedding. We use a single observation of additional pre-transition context, i.e. $k=1$. When decoding the latent policy in the online environment, we have two decoder heads on top of the latent policy whose logits are averaged per action before applying softmax. One head is trained in a supervised manner on $(a_t, z_t)$ tuples from historical data, the other is trained via Proximal Policy Optimization~\citep[PPO,][]{ppo}. We keep all convolutional layers frozen and found that a much larger learning rate of 0.01 can be stably used when only training these final few layers. We similarly tuned the learning rate for training the full network from scratch, but found no improvement compared to the original value of 5e-4. Other hyperparameters are given in Appendix \ref{apdx:hps}.

\section{Results and Discussion}
\label{sec:results}

\subsection{Decoding the latent policy online}
\label{sec:result_rlft}

\begin{figure}
    \vspace{-8mm}
     \begin{subfigure}[b]{0.64\textwidth}
         \centering
         \includegraphics[width=\textwidth]{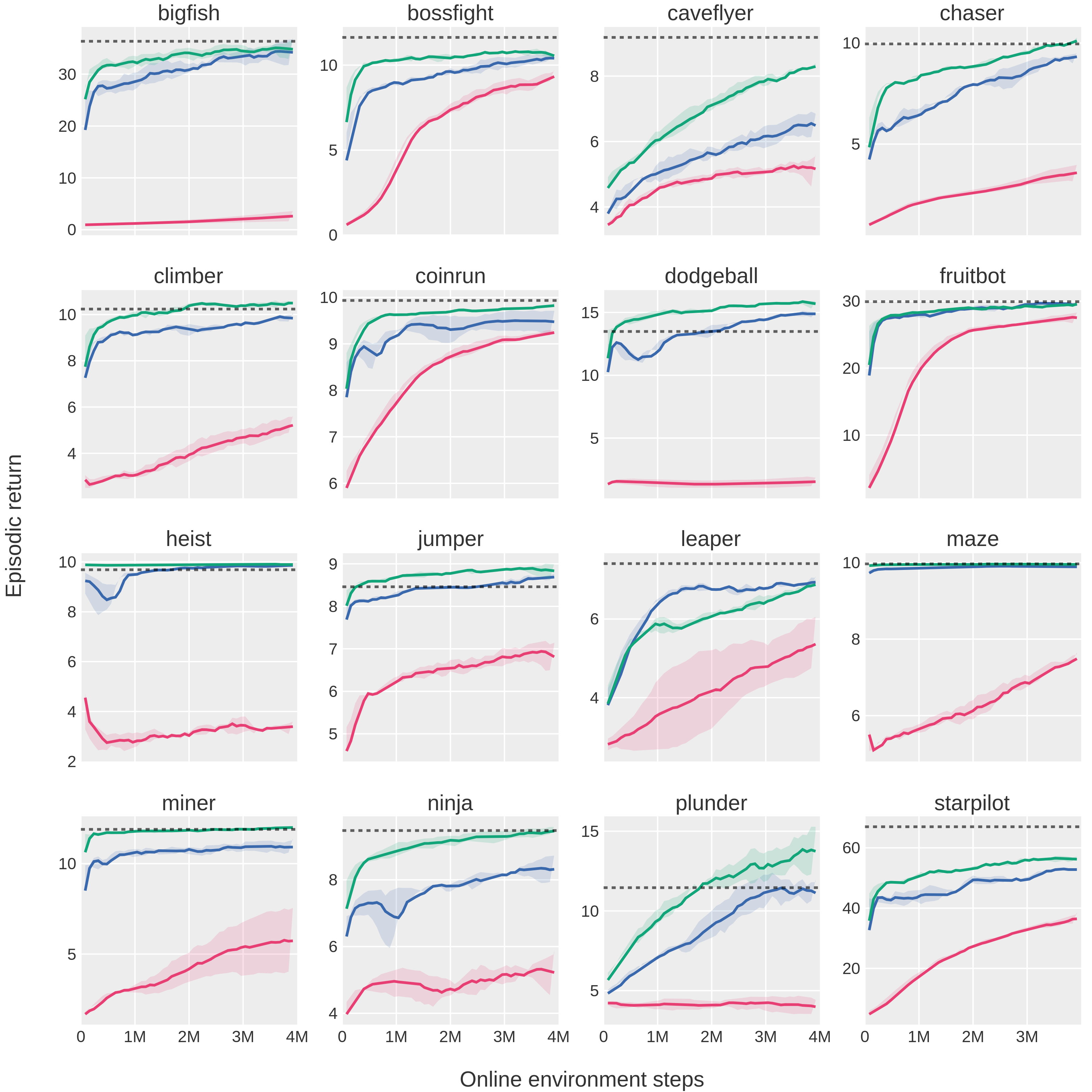}
     \end{subfigure}
     \hfill
     \begin{subfigure}[b]{0.35\textwidth}
         \centering
         \raisebox{20.5mm}{\includegraphics[width=\textwidth]{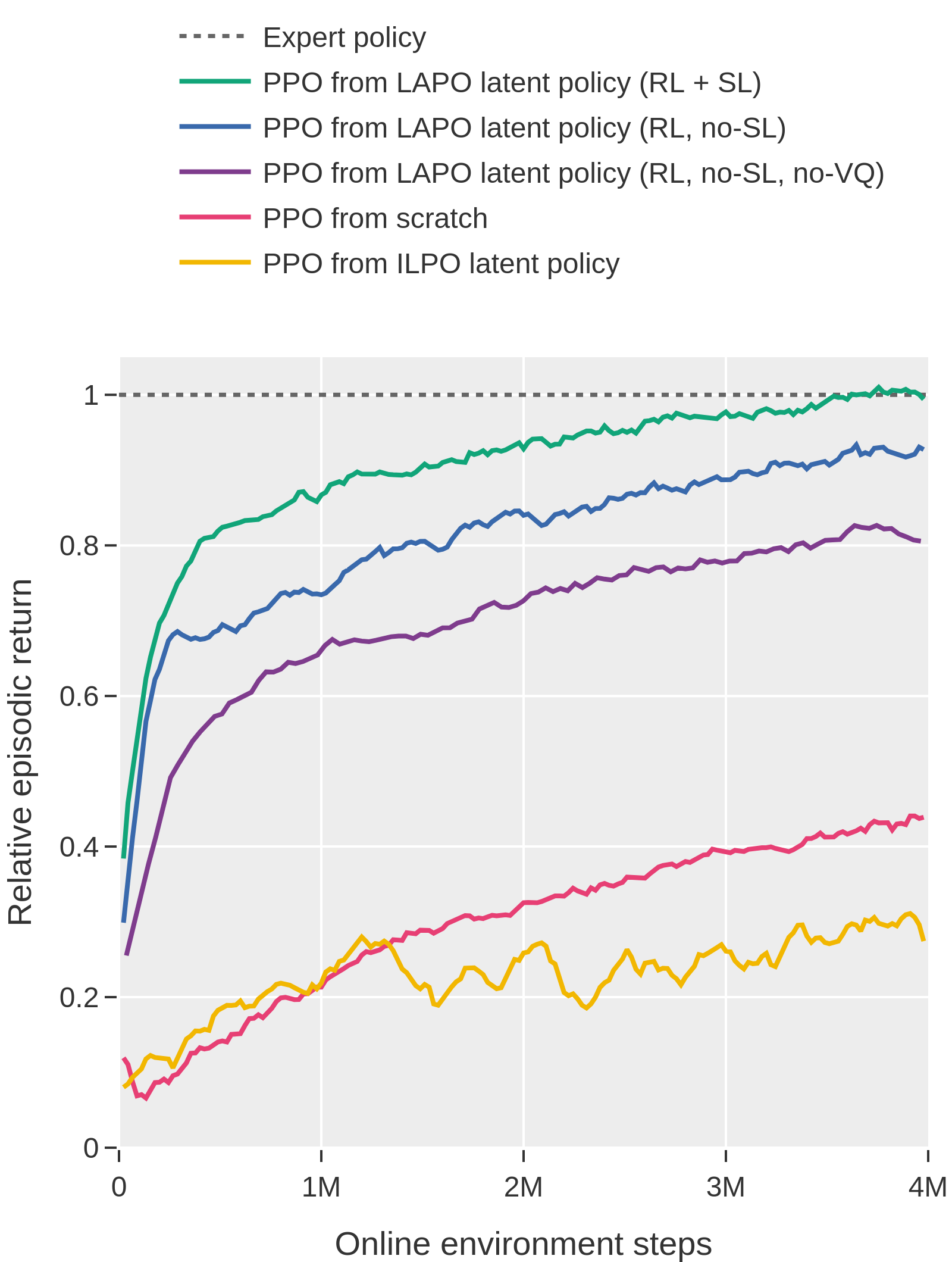}}
     \end{subfigure}
     \caption{Left: Mean episodic returns (over the course of training) for decoding \method{}'s latent policy and PPO from scratch (averaged across 3 seeds). Right: Mean test returns relative to per-environment expert policies averaged across all 16 Procgen environments. Error bars indicate standard deviation across seeds.}
     \label{fig:procgen_rlft_results}
    \vspace{-2mm}
\end{figure}

We now discuss our results for decoding the latent policy in the online environment. Recall, that this is the third stage of our approach, after first training a latent IDM, followed by obtaining a latent policy via behavior cloning. In this experiment, we compare how decoding the latent policy (via supervised learning and RL) compares to learning a policy from scratch purely from environment interaction via PPO. Since our work focuses on the setting where large amounts of observation-only data are freely available, but environment interaction is limited or expensive, we run PPO for 4M steps rather than the 25M steps suggested by \citet{procgen}. Experiments are performed separately per environment. As can be seen in Figure~\ref{fig:procgen_rlft_results}, using \method{}'s latent policy we are able to fully recover expert performance within only 4M frames, while PPO from scratch reaches only 44\% of expert performance in the same period. We highlight that in 9 of 16 tasks, our approach exceeds the performance of the expert (which was trained for 25M frames) within only 4M frames. We also provide results for two ablations. The ``RL, no-SL'' ablation trains the decoder only using reinforcement learning. In this setting, \method{} can be viewed as an RL-pretraining step that produces an initial policy representation that is useful for downstream RL. The ``RL, no-SL, no-VQ'' ablation additionally removes vector-quantization, showing that VQ is an important component of our method.

We use ILPO as our primary baseline, but found that its policy immediately collapses in several Procgen tasks. When it does not collapse, online decoding of the ILPO policy did not perform better than PPO from scratch. While \citet{ilpo} do provide positive results for \texttt{CoinRun} (which we reproduce in Appendix~\ref{apdx:ilpo_repro}), their experiments target only individual levels of this single task. We thus hypothesize that ILPO's policy collapse is in part due to the modeling challenge posed by the visually diverse levels in the full, procedurally-generated setting. By using an IDM rather than a policy, plus a high-dimensional, continuous latent space, \method{} can capture stochasticity and unobserved information (such as off-screen parts of the game). In contrast, it is likely difficult for ILPO to model the full distribution of possible transitions using only a few discrete latents.

\subsection{Decoding the latent policy offline}

\begin{wrapfigure}{r}{0.47\textwidth}
    \vspace{-5mm}
    \centering
    \includegraphics[width=0.45\textwidth]{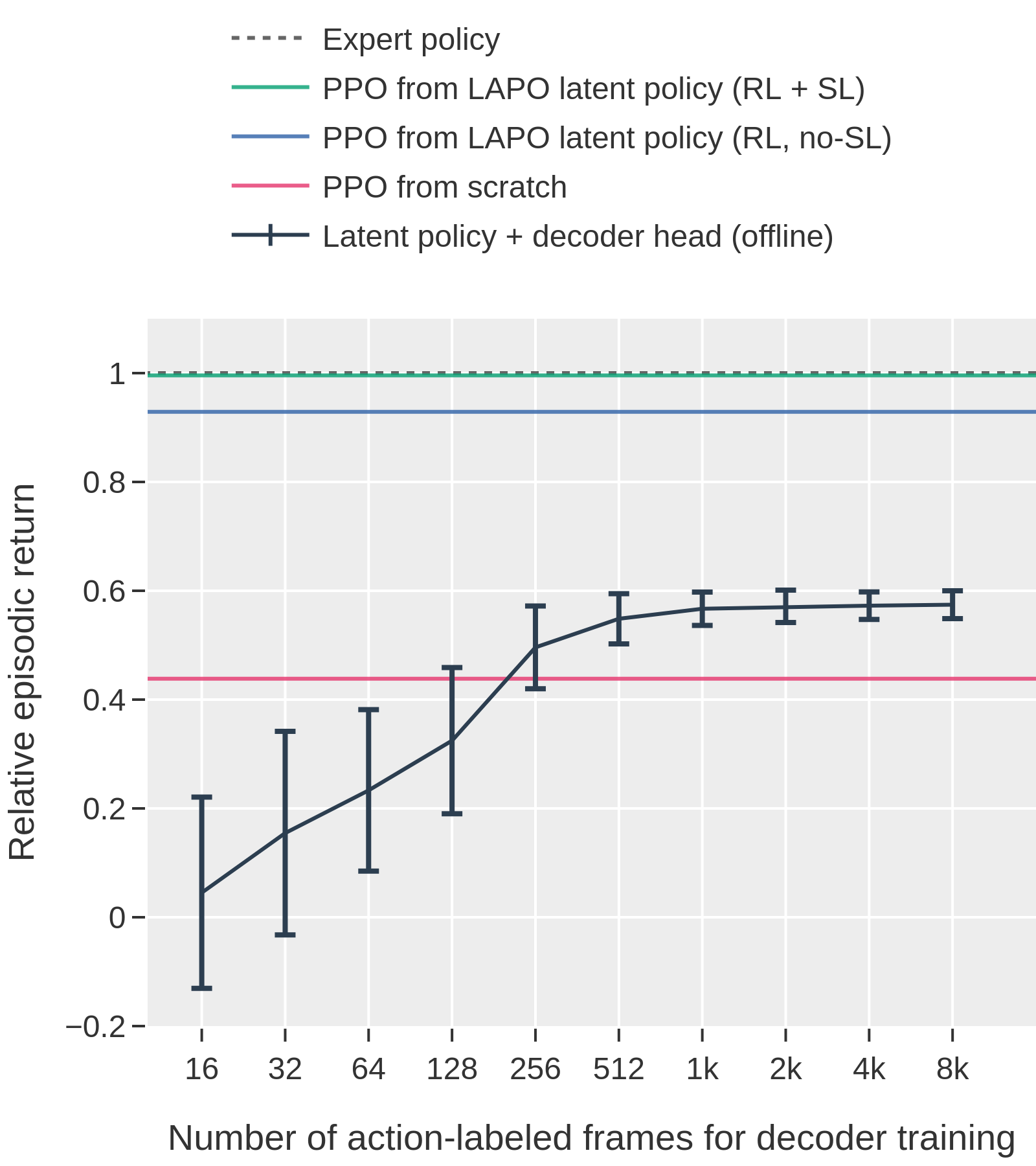}
    \caption{\small{Offline decoding performance vs. \# labeled transitions (Mean and std across 3 seeds).}}
    \vspace{-5mm}
    \label{fig:offline_decoded}
\end{wrapfigure}

Next, we consider training a latent action decoder fully-offline using only a small action-labeled dataset and no access to the online environment. The trained decoder is then composed with the latent policy to map the policy's actions into the true action space. As we will see in Section~\ref{sec:latent_space_ana}, the latent and true action spaces share similar structure. This similarity allows an effective decoder to be trained on only a miniscule amount of action-labeled data. Figure~\ref{fig:offline_decoded} shows that a decoder trained on less than 256 labeled transitions matches the performance of a policy trained from scratch for 4M steps. However, we observe that with increasing dataset size, performance eventually plateaus below the level of online decoder training. This is likely because, as previously noted, latent actions are not necessarily state-invariant. A decoder that only has access to a latent action, but not the state in which that action is being taken, may not always be able to successfully decode it. We provide per-environment results in Appendix~\ref{apdx:offline_dec_results}.

\subsection{Inspecting the Latent Action Space}
\label{sec:latent_space_ana}

\begin{figure}[t!]
\centering
\includegraphics[width=1.0\textwidth]{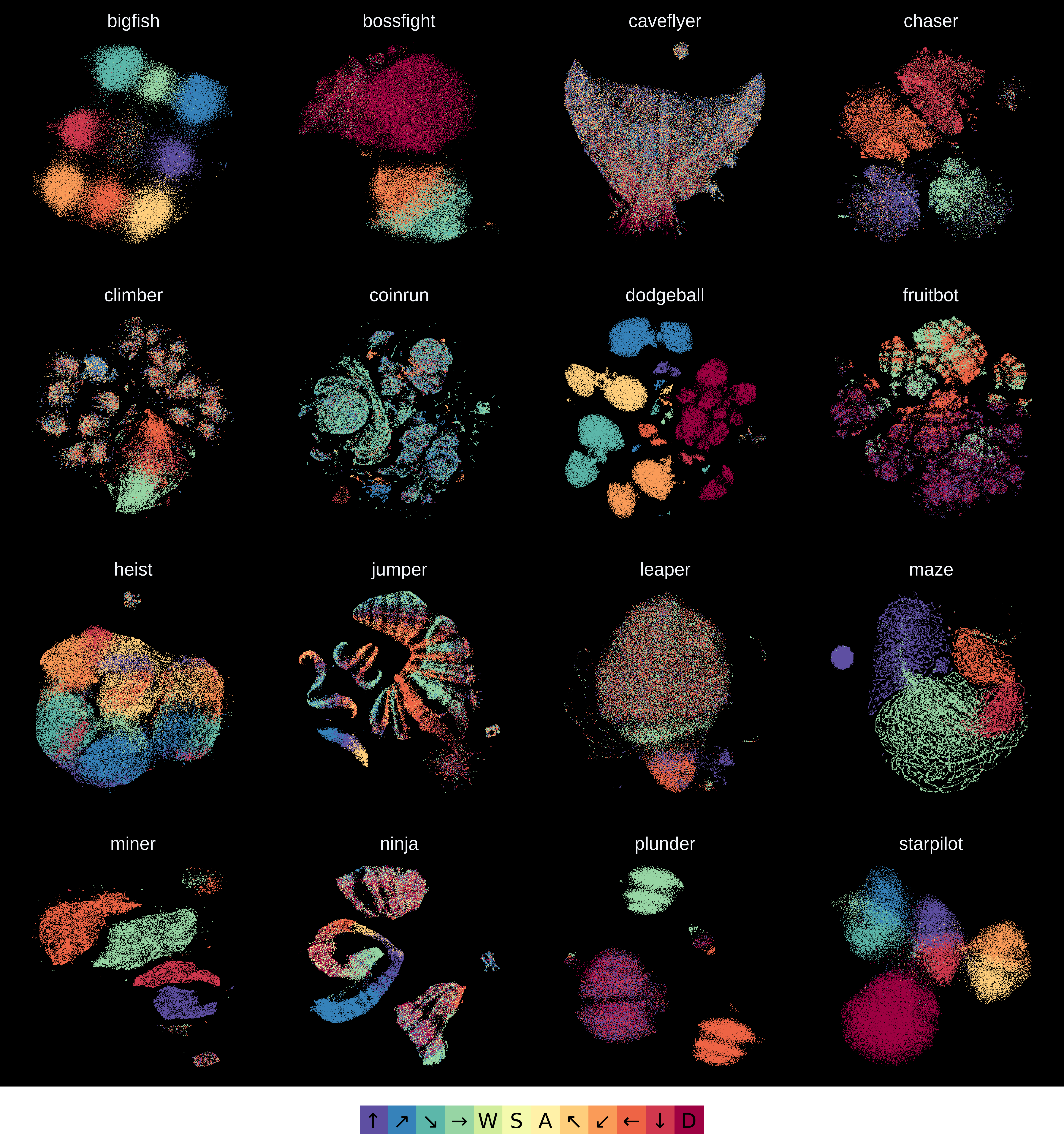}
\caption{\small{UMAP projection of the learned latent action space for all 16 procgen games. Each point represents the continuous (pre-quantization) latent action generated by the IDM for a transition in the observation-only dataset. Each point is color-coded by the true action taken by the agent at that transition (true action labels are only for visualization, not used for training). Arrows in the legend correspond to movement directions. \texttt{NOOP} actions are omitted for clarity.}}
\label{fig:umap_grid}
\end{figure}

To better understand what \method{} learns, we generate a UMAP projection~\citep{umap} of the latent action space for each of the 16 Procgen environments, shown in Figure~\ref{fig:umap_grid}. For most games, the structure of the latent action space is highly interpretable, with distinct clusters of latent actions closely corresponding to the true discrete actions. We also observe that the structure of the latent space varies greatly across environments. In \texttt{BigFish}, \texttt{Chaser}, \texttt{Dodgeball}, \texttt{Maze}, \texttt{Miner}, \texttt{Plunder}, and \texttt{StarPilot}, latent actions aggregate together in well-defined clusters aligned with the true actions. In other environments, including \texttt{Climber}, \texttt{CoinRun}, \texttt{FruitBot}, \texttt{Jumper} and \texttt{Ninja}, the latent action space exhibits more fragmented structure. This split roughly corresponds to environments with higher or lower degrees of partial observability. In the latter group of environments, the pixel observation shows only part of the level, cropped locally to the player character. Thus, when the player is moving, most observations will show parts of the level that were hidden in previous observations. In order for the FDM to accurately predict the next observation, the IDM needs to encode this off-screen information into the latent action, which we hypothesize leads to more complex structures in the latent space. However, as demonstrated in Section~\ref{sec:result_rlft}, the latent-space policy still performs well downstream even on environments with greater degrees of partial observability. Moreso, we find that vector quantization has a large impact in simplifying the structure of the latent action space. We generate the same UMAP projection for an ablation of our method without vector quantization in Appendix~\ref{apdx:latent_space_novq}. We note that for several environments, including those with and without partial observability, latent action clusters corresponding to distinct true actions are at times duplicated. For example, in \texttt{BigFish}, each true action sees four latent action clusters---likely reflecting how taking the same true action in different parts of the state-space can have differing effects. These results strongly support our hypothesis that vector quantization leads to simpler latent representations with less fragmentation within actions.

\subsection{Limitations}

A few factors can adversely impact the performance of \method{}. First, actions that have a delayed effect in observations will be predicted to take place with the same delay, i.e. the latent policy actually models the visible effects of an action, not the action itself. Nevertheless, in most environments, actions that have any impact on the state of the environment will elicit some degree of immediate change in the observation. Moreover, delayed actions can be partially addressed by extending the IDM and FDM architecture to consider multiple timesteps into the past and future, e.g. by using a Transformer-based architecture~\citep{vaswani2017attention}. Second, significant stochasticity can make it difficult for the IDM to compress the useful bits of information among the noise, degrading the quality of the latent representation. This issue can potentially be mitigated by training on much larger datasets. Lastly, training on much larger datasets---as would be required for modeling more complex domains like web-scale video---would require scaling up the model architecture, which introduces new challenges in balancing the strength of the FDM and the capacity of latent actions representations, as is often the case in autoencoding architectures~\citep{chen2016variational}.

\section{Conclusion}
\label{sec:conclusion}
This work introduced \method{}, a method for training policies over a learned latent action space, inferred from purely observational data. Unlike prior work on imitation learning from observation, \method{} does not rely on access to the true action space or a predefined set of discrete latent actions to learn a useful, pretrained policy. Instead, \method{} learns a latent action space end-to-end, by optimizing an unsupervised objective based on predictive consistency between an inverse and a forward dynamics model. Vector quantization of the continuous latent actions induces an information bottleneck that forces the quantized actions to capture state-invariant transition information. Across all 16 games of the challenging Procgen Benchmark, we show that this approach can learn latent action spaces that reflect the structure of the true action spaces, despite \method{} never having access to the latter. We then demonstrate that latent action policies, obtained through behavior cloning of latent actions, can serve as useful pretrained models that can be rapidly adapted to recover or even exceed the performance of the expert that generated the original action-free dataset.

Our results thus suggest that \method{} can serve as a useful approach for obtaining rapidly-adaptable pretrained policies from web-scale action-free demonstration data, e.g. in the form of videos. We believe \method{} could serve as an important step in unlocking the web-scale unsupervised pretraining paradigm that has proven effective in other domains like language and vision~\citep{brown2020language, radford2021learning,ramesh2021zero}. Toward this goal, we are excited about future work aimed at scaling up the world model and policy to more powerful architectures that can consider multiple timesteps and richer observations. Such models may enable efficient learning of generalist latent action policies in complex, multi-task settings, all within a single \method{} instance, while enabling more effective generalization to tasks unseen during training.

\newpage

\bibliography{refs}

\begin{thebibliography}{50}
\providecommand{\natexlab}[1]{#1}
\providecommand{\url}[1]{\texttt{#1}}
\expandafter\ifx\csname urlstyle\endcsname\relax
  \providecommand{\doi}[1]{doi: #1}\else
  \providecommand{\doi}{doi: \begingroup \urlstyle{rm}\Url}\fi

\bibitem[{\AA}str{\"o}m(1965)]{aastrom1965optimal}
Karl~Johan {\AA}str{\"o}m.
\newblock Optimal control of {Markov} processes with incomplete state information.
\newblock \emph{Journal of Mathematical Analysis and Applications}, 10\penalty0 (1):\penalty0 174--205, 1965.

\bibitem[Aytar et~al.(2018)Aytar, Pfaff, Budden, Paine, Wang, and de~Freitas]{hard_expl_youtube}
Yusuf Aytar, Tobias Pfaff, David Budden, Tom~Le Paine, Ziyu Wang, and Nando de~Freitas.
\newblock Playing hard exploration games by watching youtube.
\newblock In Samy Bengio, Hanna~M. Wallach, Hugo Larochelle, Kristen Grauman, Nicol{\`{o}} Cesa{-}Bianchi, and Roman Garnett (eds.), \emph{Advances in Neural Information Processing Systems 31: Annual Conference on Neural Information Processing Systems 2018, NeurIPS 2018, December 3-8, 2018, Montr{\'{e}}al, Canada}, pp.\  2935--2945, 2018.
\newblock URL \url{https://proceedings.neurips.cc/paper/2018/hash/35309226eb45ec366ca86a4329a2b7c3-Abstract.html}.

\bibitem[Baker et~al.(2022)Baker, Akkaya, Zhokov, Huizinga, Tang, Ecoffet, Houghton, Sampedro, and Clune]{vpt}
Bowen Baker, Ilge Akkaya, Peter Zhokov, Joost Huizinga, Jie Tang, Adrien Ecoffet, Brandon Houghton, Raul Sampedro, and Jeff Clune.
\newblock Video pretraining {(VPT):} learning to act by watching unlabeled online videos.
\newblock In \emph{NeurIPS}, 2022.
\newblock URL \url{http://papers.nips.cc/paper\_files/paper/2022/hash/9c7008aff45b5d8f0973b23e1a22ada0-Abstract-Conference.html}.

\bibitem[Bellemare et~al.(2013)Bellemare, Naddaf, Veness, and Bowling]{bellemare2013arcade}
Marc~G Bellemare, Yavar Naddaf, Joel Veness, and Michael Bowling.
\newblock The arcade learning environment: An evaluation platform for general agents.
\newblock \emph{Journal of Artificial Intelligence Research}, 47:\penalty0 253--279, 2013.

\bibitem[Bengio et~al.(2013)Bengio, L{\'{e}}onard, and Courville]{straight_through_est}
Yoshua Bengio, Nicholas L{\'{e}}onard, and Aaron~C. Courville.
\newblock Estimating or propagating gradients through stochastic neurons for conditional computation.
\newblock \emph{CoRR}, abs/1308.3432, 2013.
\newblock URL \url{http://arxiv.org/abs/1308.3432}.

\bibitem[Brown et~al.(2020)Brown, Mann, Ryder, Subbiah, Kaplan, Dhariwal, Neelakantan, Shyam, Sastry, Askell, et~al.]{brown2020language}
Tom Brown, Benjamin Mann, Nick Ryder, Melanie Subbiah, Jared~D Kaplan, Prafulla Dhariwal, Arvind Neelakantan, Pranav Shyam, Girish Sastry, Amanda Askell, et~al.
\newblock Language models are few-shot learners.
\newblock \emph{Advances in neural information processing systems}, 33:\penalty0 1877--1901, 2020.

\bibitem[Caron et~al.(2021)Caron, Touvron, Misra, J{\'{e}}gou, Mairal, Bojanowski, and Joulin]{dino}
Mathilde Caron, Hugo Touvron, Ishan Misra, Herv{\'{e}} J{\'{e}}gou, Julien Mairal, Piotr Bojanowski, and Armand Joulin.
\newblock Emerging properties in self-supervised vision transformers.
\newblock In \emph{2021 {IEEE/CVF} International Conference on Computer Vision, {ICCV} 2021, Montreal, QC, Canada, October 10-17, 2021}, pp.\  9630--9640. {IEEE}, 2021.
\newblock \doi{10.1109/ICCV48922.2021.00951}.
\newblock URL \url{https://doi.org/10.1109/ICCV48922.2021.00951}.

\bibitem[Chen et~al.(2016)Chen, Kingma, Salimans, Duan, Dhariwal, Schulman, Sutskever, and Abbeel]{chen2016variational}
Xi~Chen, Diederik~P Kingma, Tim Salimans, Yan Duan, Prafulla Dhariwal, John Schulman, Ilya Sutskever, and Pieter Abbeel.
\newblock Variational lossy autoencoder.
\newblock In \emph{International Conference on Learning Representations}, 2016.

\bibitem[Cobbe et~al.(2019)Cobbe, Klimov, Hesse, Kim, and Schulman]{cobbe2019quantifying}
Karl Cobbe, Oleg Klimov, Chris Hesse, Taehoon Kim, and John Schulman.
\newblock Quantifying generalization in reinforcement learning.
\newblock In \emph{International Conference on Machine Learning}, pp.\  1282--1289. PMLR, 2019.

\bibitem[Cobbe et~al.(2020)Cobbe, Hesse, Hilton, and Schulman]{procgen}
Karl Cobbe, Christopher Hesse, Jacob Hilton, and John Schulman.
\newblock Leveraging procedural generation to benchmark reinforcement learning.
\newblock In \emph{Proceedings of the 37th International Conference on Machine Learning, {ICML} 2020, 13-18 July 2020, Virtual Event}, volume 119 of \emph{Proceedings of Machine Learning Research}, pp.\  2048--2056. {PMLR}, 2020.
\newblock URL \url{http://proceedings.mlr.press/v119/cobbe20a.html}.

\bibitem[Dhariwal et~al.(2020)Dhariwal, Jun, Payne, Kim, Radford, and Sutskever]{vq_audio_jukebox}
Prafulla Dhariwal, Heewoo Jun, Christine Payne, Jong~Wook Kim, Alec Radford, and Ilya Sutskever.
\newblock Jukebox: {A} generative model for music.
\newblock \emph{CoRR}, abs/2005.00341, 2020.
\newblock URL \url{https://arxiv.org/abs/2005.00341}.

\bibitem[Edwards et~al.(2019)Edwards, Sahni, Schroecker, and Jr.]{ilpo}
Ashley~D. Edwards, Himanshu Sahni, Yannick Schroecker, and Charles L.~Isbell Jr.
\newblock Imitating latent policies from observation.
\newblock In Kamalika Chaudhuri and Ruslan Salakhutdinov (eds.), \emph{Proceedings of the 36th International Conference on Machine Learning, {ICML} 2019, 9-15 June 2019, Long Beach, California, {USA}}, volume~97 of \emph{Proceedings of Machine Learning Research}, pp.\  1755--1763. {PMLR}, 2019.
\newblock URL \url{http://proceedings.mlr.press/v97/edwards19a.html}.

\bibitem[Espeholt et~al.(2018)Espeholt, Soyer, Munos, Simonyan, Mnih, Ward, Doron, Firoiu, Harley, Dunning, Legg, and Kavukcuoglu]{impala}
Lasse Espeholt, Hubert Soyer, R{\'{e}}mi Munos, Karen Simonyan, Volodymyr Mnih, Tom Ward, Yotam Doron, Vlad Firoiu, Tim Harley, Iain Dunning, Shane Legg, and Koray Kavukcuoglu.
\newblock {IMPALA:} scalable distributed deep-rl with importance weighted actor-learner architectures.
\newblock In Jennifer~G. Dy and Andreas Krause (eds.), \emph{Proceedings of the 35th International Conference on Machine Learning, {ICML} 2018, Stockholmsm{\"{a}}ssan, Stockholm, Sweden, July 10-15, 2018}, volume~80 of \emph{Proceedings of Machine Learning Research}, pp.\  1406--1415. {PMLR}, 2018.
\newblock URL \url{http://proceedings.mlr.press/v80/espeholt18a.html}.

\bibitem[Ghosh et~al.(2023)Ghosh, Bhateja, and Levine]{icvf}
Dibya Ghosh, Chethan~Anand Bhateja, and Sergey Levine.
\newblock Reinforcement learning from passive data via latent intentions.
\newblock In Andreas Krause, Emma Brunskill, Kyunghyun Cho, Barbara Engelhardt, Sivan Sabato, and Jonathan Scarlett (eds.), \emph{International Conference on Machine Learning, {ICML} 2023, 23-29 July 2023, Honolulu, Hawaii, {USA}}, volume 202 of \emph{Proceedings of Machine Learning Research}, pp.\  11321--11339. {PMLR}, 2023.
\newblock URL \url{https://proceedings.mlr.press/v202/ghosh23a.html}.

\bibitem[Gray(1984)]{vq}
R.~Gray.
\newblock Vector quantization.
\newblock \emph{IEEE ASSP Magazine}, 1\penalty0 (2):\penalty0 4--29, 1984.
\newblock \doi{10.1109/MASSP.1984.1162229}.

\bibitem[He et~al.(2016)He, Zhang, Ren, and Sun]{resnet}
Kaiming He, Xiangyu Zhang, Shaoqing Ren, and Jian Sun.
\newblock Deep residual learning for image recognition.
\newblock In \emph{Proceedings of the IEEE conference on computer vision and pattern recognition}, pp.\  770--778, 2016.

\bibitem[Huh et~al.(2023)Huh, Cheung, Agrawal, and Isola]{straightening_out_vq}
Minyoung Huh, Brian Cheung, Pulkit Agrawal, and Phillip Isola.
\newblock Straightening out the straight-through estimator: Overcoming optimization challenges in vector quantized networks.
\newblock In Andreas Krause, Emma Brunskill, Kyunghyun Cho, Barbara Engelhardt, Sivan Sabato, and Jonathan Scarlett (eds.), \emph{International Conference on Machine Learning, {ICML} 2023, 23-29 July 2023, Honolulu, Hawaii, {USA}}, volume 202 of \emph{Proceedings of Machine Learning Research}, pp.\  14096--14113. {PMLR}, 2023.
\newblock URL \url{https://proceedings.mlr.press/v202/huh23a.html}.

\bibitem[Hutter(2003)]{hutter2003existence}
Marcus Hutter.
\newblock On the existence and convergence of computable universal priors.
\newblock In \emph{International Conference on Algorithmic Learning Theory}, pp.\  298--312. Springer, 2003.

\bibitem[Kaelbling et~al.(1998)Kaelbling, Littman, and Cassandra]{kaelbling1998planning}
Leslie~Pack Kaelbling, Michael~L Littman, and Anthony~R Cassandra.
\newblock Planning and acting in partially observable stochastic domains.
\newblock \emph{Artificial intelligence}, 101\penalty0 (1-2):\penalty0 99--134, 1998.

\bibitem[Kumar et~al.(2020)Kumar, Zhou, Tucker, and Levine]{kumar2020conservative}
Aviral Kumar, Aurick Zhou, George Tucker, and Sergey Levine.
\newblock Conservative q-learning for offline reinforcement learning.
\newblock \emph{Advances in Neural Information Processing Systems}, 33:\penalty0 1179--1191, 2020.

\bibitem[Levine et~al.(2020)Levine, Kumar, Tucker, and Fu]{levine2020offline}
Sergey Levine, Aviral Kumar, George Tucker, and Justin Fu.
\newblock Offline reinforcement learning: Tutorial, review, and perspectives on open problems.
\newblock \emph{arXiv preprint arXiv:2005.01643}, 2020.

\bibitem[McInnes et~al.(2018)McInnes, Healy, and Melville]{umap}
Leland McInnes, John Healy, and James Melville.
\newblock Umap: Uniform manifold approximation and projection for dimension reduction.
\newblock \emph{arXiv preprint arXiv:1802.03426}, 2018.

\bibitem[Menapace et~al.(2021)Menapace, Lathuili{\`{e}}re, Tulyakov, Siarohin, and Ricci]{pvg}
Willi Menapace, St{\'{e}}phane Lathuili{\`{e}}re, Sergey Tulyakov, Aliaksandr Siarohin, and Elisa Ricci.
\newblock Playable video generation.
\newblock In \emph{{IEEE} Conference on Computer Vision and Pattern Recognition, {CVPR} 2021, virtual, June 19-25, 2021}, pp.\  10061--10070. Computer Vision Foundation / {IEEE}, 2021.
\newblock \doi{10.1109/CVPR46437.2021.00993}.
\newblock URL \url{https://openaccess.thecvf.com/content/CVPR2021/html/Menapace\_Playable\_Video\_Generation\_CVPR\_2021\_paper.html}.

\bibitem[Micheli et~al.(2023)Micheli, Alonso, and Fleuret]{iris}
Vincent Micheli, Eloi Alonso, and Fran{\c{c}}ois Fleuret.
\newblock Transformers are sample-efficient world models.
\newblock In \emph{The Eleventh International Conference on Learning Representations, {ICLR} 2023, Kigali, Rwanda, May 1-5, 2023}. OpenReview.net, 2023.
\newblock URL \url{https://openreview.net/pdf?id=vhFu1Acb0xb}.

\bibitem[Polyak \& Juditsky(1992)Polyak and Juditsky]{polyak1992acceleration}
Boris~T Polyak and Anatoli~B Juditsky.
\newblock Acceleration of stochastic approximation by averaging.
\newblock \emph{SIAM journal on control and optimization}, 30\penalty0 (4):\penalty0 838--855, 1992.

\bibitem[Pomerleau(1988)]{pomerleau1988alvinn}
Dean Pomerleau.
\newblock {ALVINN:} an autonomous land vehicle in a neural network.
\newblock In David~S. Touretzky (ed.), \emph{Advances in Neural Information Processing Systems 1, {[NIPS} Conference, Denver, Colorado, USA, 1988]}, pp.\  305--313. Morgan Kaufmann, 1988.

\bibitem[Radford et~al.(2019)Radford, Wu, Child, Luan, Amodei, and Sutskever]{gpt2}
Alec Radford, Jeff Wu, Rewon Child, David Luan, Dario Amodei, and Ilya Sutskever.
\newblock Language models are unsupervised multitask learners.
\newblock 2019.

\bibitem[Radford et~al.(2021)Radford, Kim, Hallacy, Ramesh, Goh, Agarwal, Sastry, Askell, Mishkin, Clark, et~al.]{radford2021learning}
Alec Radford, Jong~Wook Kim, Chris Hallacy, Aditya Ramesh, Gabriel Goh, Sandhini Agarwal, Girish Sastry, Amanda Askell, Pamela Mishkin, Jack Clark, et~al.
\newblock Learning transferable visual models from natural language supervision.
\newblock In \emph{International conference on machine learning}, pp.\  8748--8763. PMLR, 2021.

\bibitem[Ramesh et~al.(2021{\natexlab{a}})Ramesh, Pavlov, Goh, Gray, Voss, Radford, Chen, and Sutskever]{ramesh2021zero}
Aditya Ramesh, Mikhail Pavlov, Gabriel Goh, Scott Gray, Chelsea Voss, Alec Radford, Mark Chen, and Ilya Sutskever.
\newblock Zero-shot text-to-image generation.
\newblock In \emph{International Conference on Machine Learning}, pp.\  8821--8831. PMLR, 2021{\natexlab{a}}.

\bibitem[Ramesh et~al.(2021{\natexlab{b}})Ramesh, Pavlov, Goh, Gray, Voss, Radford, Chen, and Sutskever]{vq_for_img}
Aditya Ramesh, Mikhail Pavlov, Gabriel Goh, Scott Gray, Chelsea Voss, Alec Radford, Mark Chen, and Ilya Sutskever.
\newblock Zero-shot text-to-image generation.
\newblock In Marina Meila and Tong Zhang (eds.), \emph{Proceedings of the 38th International Conference on Machine Learning, {ICML} 2021, 18-24 July 2021, Virtual Event}, volume 139 of \emph{Proceedings of Machine Learning Research}, pp.\  8821--8831. {PMLR}, 2021{\natexlab{b}}.
\newblock URL \url{http://proceedings.mlr.press/v139/ramesh21a.html}.

\bibitem[Reed et~al.(2022)Reed, Zolna, Parisotto, Colmenarejo, Novikov, Barth{-}Maron, Gimenez, Sulsky, Kay, Springenberg, Eccles, Bruce, Razavi, Edwards, Heess, Hadsell, Vinyals, Bordbar, and de~Freitas]{gato}
Scott~E. Reed, Konrad Zolna, Emilio Parisotto, Sergio~G{\'{o}}mez Colmenarejo, Alexander Novikov, Gabriel Barth{-}Maron, Mai Gimenez, Yury Sulsky, Jackie Kay, Jost~Tobias Springenberg, Tom Eccles, Jake Bruce, Ali Razavi, Ashley Edwards, Nicolas Heess, Raia Hadsell, Oriol Vinyals, Mahyar Bordbar, and Nando de~Freitas.
\newblock A generalist agent.
\newblock \emph{Trans. Mach. Learn. Res.}, 2022, 2022.
\newblock URL \url{https://openreview.net/forum?id=1ikK0kHjvj}.

\bibitem[Ronneberger et~al.(2015)Ronneberger, Fischer, and Brox]{ronneberger2015unet}
Olaf Ronneberger, Philipp Fischer, and Thomas Brox.
\newblock U-net: Convolutional networks for biomedical image segmentation.
\newblock In \emph{Medical Image Computing and Computer-Assisted Intervention--MICCAI 2015: 18th International Conference, Munich, Germany, October 5-9, 2015, Proceedings, Part III 18}, pp.\  234--241. Springer, 2015.

\bibitem[Ross \& Bagnell(2010)Ross and Bagnell]{ross2010efficient}
St{\'e}phane Ross and Drew Bagnell.
\newblock Efficient reductions for imitation learning.
\newblock In \emph{Proceedings of the thirteenth international conference on artificial intelligence and statistics}, pp.\  661--668. JMLR Workshop and Conference Proceedings, 2010.

\bibitem[Schmeckpeper et~al.(2020)Schmeckpeper, Rybkin, Daniilidis, Levine, and Finn]{rl_with_videos_idmtransfer_schmeck}
Karl Schmeckpeper, Oleh Rybkin, Kostas Daniilidis, Sergey Levine, and Chelsea Finn.
\newblock Reinforcement learning with videos: Combining offline observations with interaction.
\newblock In Jens Kober, Fabio Ramos, and Claire~J. Tomlin (eds.), \emph{4th Conference on Robot Learning, CoRL 2020, 16-18 November 2020, Virtual Event / Cambridge, MA, {USA}}, volume 155 of \emph{Proceedings of Machine Learning Research}, pp.\  339--354. {PMLR}, 2020.
\newblock URL \url{https://proceedings.mlr.press/v155/schmeckpeper21a.html}.

\bibitem[Schmidhuber(1997)]{schmidhuber1997discovering}
J{\"u}rgen Schmidhuber.
\newblock Discovering neural nets with low kolmogorov complexity and high generalization capability.
\newblock \emph{Neural Networks}, 10\penalty0 (5):\penalty0 857--873, 1997.

\bibitem[Schulman et~al.(2017)Schulman, Wolski, Dhariwal, Radford, and Klimov]{ppo}
John Schulman, Filip Wolski, Prafulla Dhariwal, Alec Radford, and Oleg Klimov.
\newblock Proximal policy optimization algorithms.
\newblock \emph{CoRR}, abs/1707.06347, 2017.
\newblock URL \url{http://arxiv.org/abs/1707.06347}.

\bibitem[Solmonoff(1964)]{solmonoff1964formal}
RJ~Solmonoff.
\newblock A formal theory of inductive inference. i.
\newblock \emph{II Information and Control}, 7:\penalty0 224--254, 1964.

\bibitem[Struckmeier \& Kyrki(2023)Struckmeier and Kyrki]{ilpo_followup}
Oliver Struckmeier and Ville Kyrki.
\newblock Preventing mode collapse when imitating latent policies from observations, 2023.
\newblock URL \url{https://openreview.net/forum?id=Mf9fQ0OgMzo}.

\bibitem[Sutton \& Barto(2018)Sutton and Barto]{sutton2018reinforcement}
Richard~S Sutton and Andrew~G Barto.
\newblock \emph{Reinforcement learning: An introduction}.
\newblock MIT press, 2018.

\bibitem[Tishby et~al.(2000)Tishby, Pereira, and Bialek]{tishby2000information}
Naftali Tishby, Fernando~C Pereira, and William Bialek.
\newblock The information bottleneck method.
\newblock \emph{arXiv preprint physics/0004057}, 2000.

\bibitem[Torabi et~al.(2018)Torabi, Warnell, and Stone]{bco}
Faraz Torabi, Garrett Warnell, and Peter Stone.
\newblock Behavioral cloning from observation.
\newblock In J{\'{e}}r{\^{o}}me Lang (ed.), \emph{Proceedings of the Twenty-Seventh International Joint Conference on Artificial Intelligence, {IJCAI} 2018, July 13-19, 2018, Stockholm, Sweden}, pp.\  4950--4957. ijcai.org, 2018.
\newblock \doi{10.24963/ijcai.2018/687}.
\newblock URL \url{https://doi.org/10.24963/ijcai.2018/687}.

\bibitem[Torabi et~al.(2019{\natexlab{a}})Torabi, Warnell, and Stone]{ifo_survey}
Faraz Torabi, Garrett Warnell, and Peter Stone.
\newblock Recent advances in imitation learning from observation.
\newblock In Sarit Kraus (ed.), \emph{Proceedings of the Twenty-Eighth International Joint Conference on Artificial Intelligence, {IJCAI} 2019, Macao, China, August 10-16, 2019}, pp.\  6325--6331. ijcai.org, 2019{\natexlab{a}}.
\newblock \doi{10.24963/ijcai.2019/882}.
\newblock URL \url{https://doi.org/10.24963/ijcai.2019/882}.

\bibitem[Torabi et~al.(2019{\natexlab{b}})Torabi, Warnell, and Stone]{torabi2018generative}
Faraz Torabi, Garrett Warnell, and Peter Stone.
\newblock Generative adversarial imitation from observation.
\newblock \emph{ICML Workshop on Imitation, Intent, and Interaction (I3)}, 2019{\natexlab{b}}.

\bibitem[van~den Oord et~al.(2017)van~den Oord, Vinyals, and Kavukcuoglu]{vqvae}
A{\"{a}}ron van~den Oord, Oriol Vinyals, and Koray Kavukcuoglu.
\newblock Neural discrete representation learning.
\newblock In Isabelle Guyon, Ulrike von Luxburg, Samy Bengio, Hanna~M. Wallach, Rob Fergus, S.~V.~N. Vishwanathan, and Roman Garnett (eds.), \emph{Advances in Neural Information Processing Systems 30: Annual Conference on Neural Information Processing Systems 2017, December 4-9, 2017, Long Beach, CA, {USA}}, pp.\  6306--6315, 2017.
\newblock URL \url{https://proceedings.neurips.cc/paper/2017/hash/7a98af17e63a0ac09ce2e96d03992fbc-Abstract.html}.

\bibitem[Vaswani et~al.(2017)Vaswani, Shazeer, Parmar, Uszkoreit, Jones, Gomez, Kaiser, and Polosukhin]{vaswani2017attention}
Ashish Vaswani, Noam Shazeer, Niki Parmar, Jakob Uszkoreit, Llion Jones, Aidan~N Gomez, {\L}ukasz Kaiser, and Illia Polosukhin.
\newblock Attention is all you need.
\newblock \emph{Advances in neural information processing systems}, 30, 2017.

\bibitem[Yang et~al.(2019)Yang, Ma, Huang, Sun, Liu, Huang, and Gan]{yang2019imitation}
Chao Yang, Xiaojian Ma, Wenbing Huang, Fuchun Sun, Huaping Liu, Junzhou Huang, and Chuang Gan.
\newblock Imitation learning from observations by minimizing inverse dynamics disagreement.
\newblock \emph{Advances in neural information processing systems}, 32, 2019.

\bibitem[Ye et~al.(2023)Ye, Zhang, Abbeel, and Gao]{latent_actions_for_mbrl}
Weirui Ye, Yunsheng Zhang, Pieter Abbeel, and Yang Gao.
\newblock Become a proficient player with limited data through watching pure videos.
\newblock In \emph{The Eleventh International Conference on Learning Representations, {ICLR} 2023, Kigali, Rwanda, May 1-5, 2023}. OpenReview.net, 2023.
\newblock URL \url{https://openreview.net/pdf?id=Sy-o2N0hF4f}.

\bibitem[Zeghidour et~al.(2022)Zeghidour, Luebs, Omran, Skoglund, and Tagliasacchi]{vq_soundstream}
Neil Zeghidour, Alejandro Luebs, Ahmed Omran, Jan Skoglund, and Marco Tagliasacchi.
\newblock Soundstream: An end-to-end neural audio codec.
\newblock \emph{{IEEE} {ACM} Trans. Audio Speech Lang. Process.}, 30:\penalty0 495--507, 2022.
\newblock \doi{10.1109/TASLP.2021.3129994}.
\newblock URL \url{https://doi.org/10.1109/TASLP.2021.3129994}.

\bibitem[Zhang et~al.(2022)Zhang, Peng, and Zhou]{DBLP:conf/eccv/ZhangPZ22}
Qihang Zhang, Zhenghao Peng, and Bolei Zhou.
\newblock Learning to drive by watching youtube videos: Action-conditioned contrastive policy pretraining.
\newblock In Shai Avidan, Gabriel~J. Brostow, Moustapha Ciss{\'{e}}, Giovanni~Maria Farinella, and Tal Hassner (eds.), \emph{Computer Vision - {ECCV} 2022 - 17th European Conference, Tel Aviv, Israel, October 23-27, 2022, Proceedings, Part {XXVI}}, volume 13686 of \emph{Lecture Notes in Computer Science}, pp.\  111--128. Springer, 2022.
\newblock \doi{10.1007/978-3-031-19809-0\_7}.
\newblock URL \url{https://doi.org/10.1007/978-3-031-19809-0\_7}.

\bibitem[Zheng et~al.(2023)Zheng, Henaff, Amos, and Grover]{ssorl}
Qinqing Zheng, Mikael Henaff, Brandon Amos, and Aditya Grover.
\newblock Semi-supervised offline reinforcement learning with action-free trajectories.
\newblock In Andreas Krause, Emma Brunskill, Kyunghyun Cho, Barbara Engelhardt, Sivan Sabato, and Jonathan Scarlett (eds.), \emph{International Conference on Machine Learning, {ICML} 2023, 23-29 July 2023, Honolulu, Hawaii, {USA}}, volume 202 of \emph{Proceedings of Machine Learning Research}, pp.\  42339--42362. {PMLR}, 2023.
\newblock URL \url{https://proceedings.mlr.press/v202/zheng23b.html}.

\end{thebibliography}
\bibliographystyle{iclr2024_conference}

\newpage
\appendix
\section{Appendix}

\subsection{Offline decoding results}
\label{apdx:offline_dec_results}

\begin{figure}[h]
\begin{center}
\includegraphics[width=1.0\textwidth]{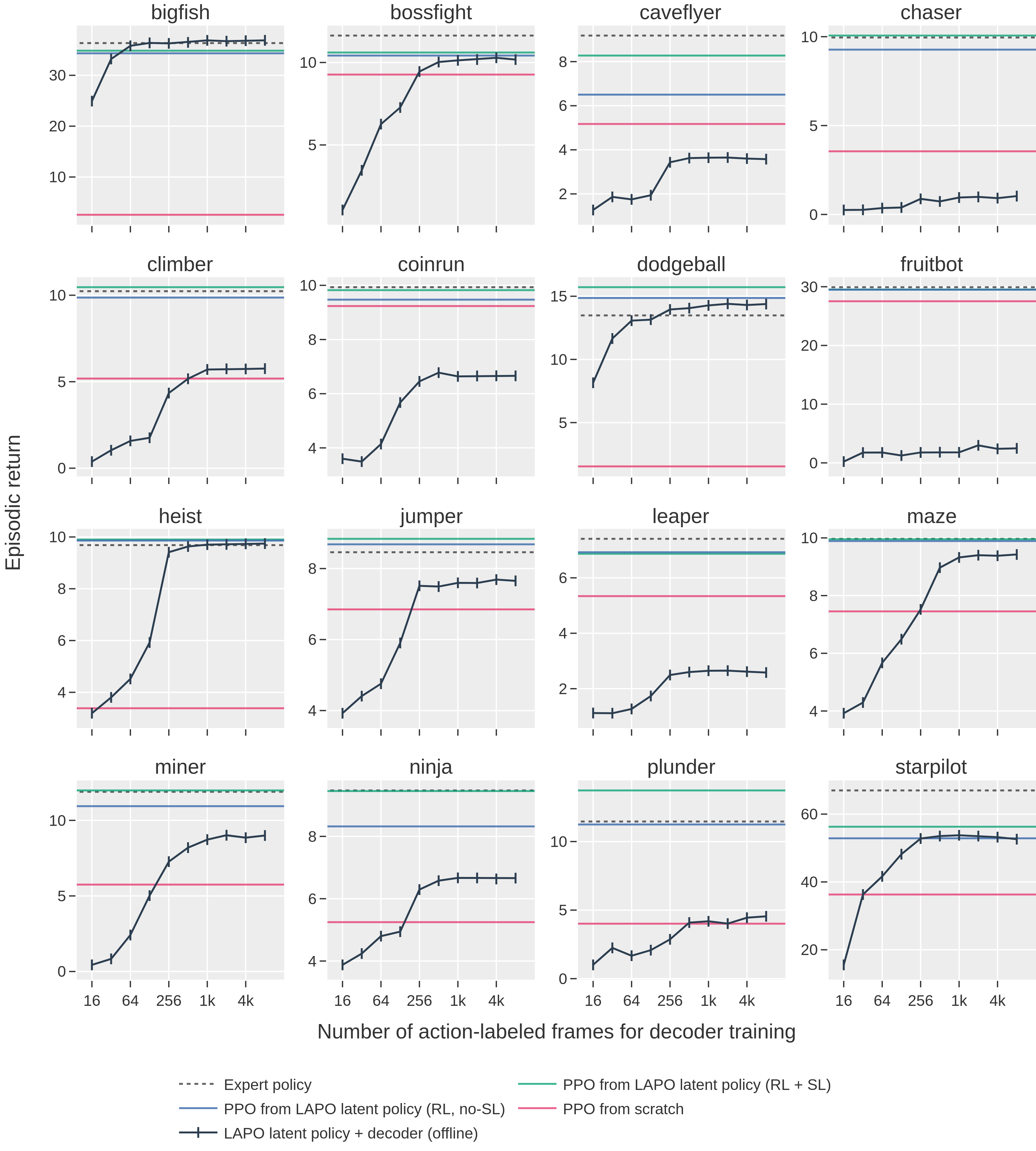}
\end{center}
\caption{Test performance of the latent policy combined with a latent action decoder that is trained on an action-labelled offline dataset of a certain size, consisting of $(z_t, a_t)$ tuples. An effective decoder can be trained with only a few hundred samples although performance generally plateaus before reaching the performance of our proposed online (RL) decoding approach.}
\label{fig:offline_decoded_grid}
\end{figure}

\newpage
\subsection{Latent space analysis for no-VQ ablation}
\label{apdx:latent_space_novq}

\begin{figure}[h]
\begin{center}
\includegraphics[width=1.0\textwidth]{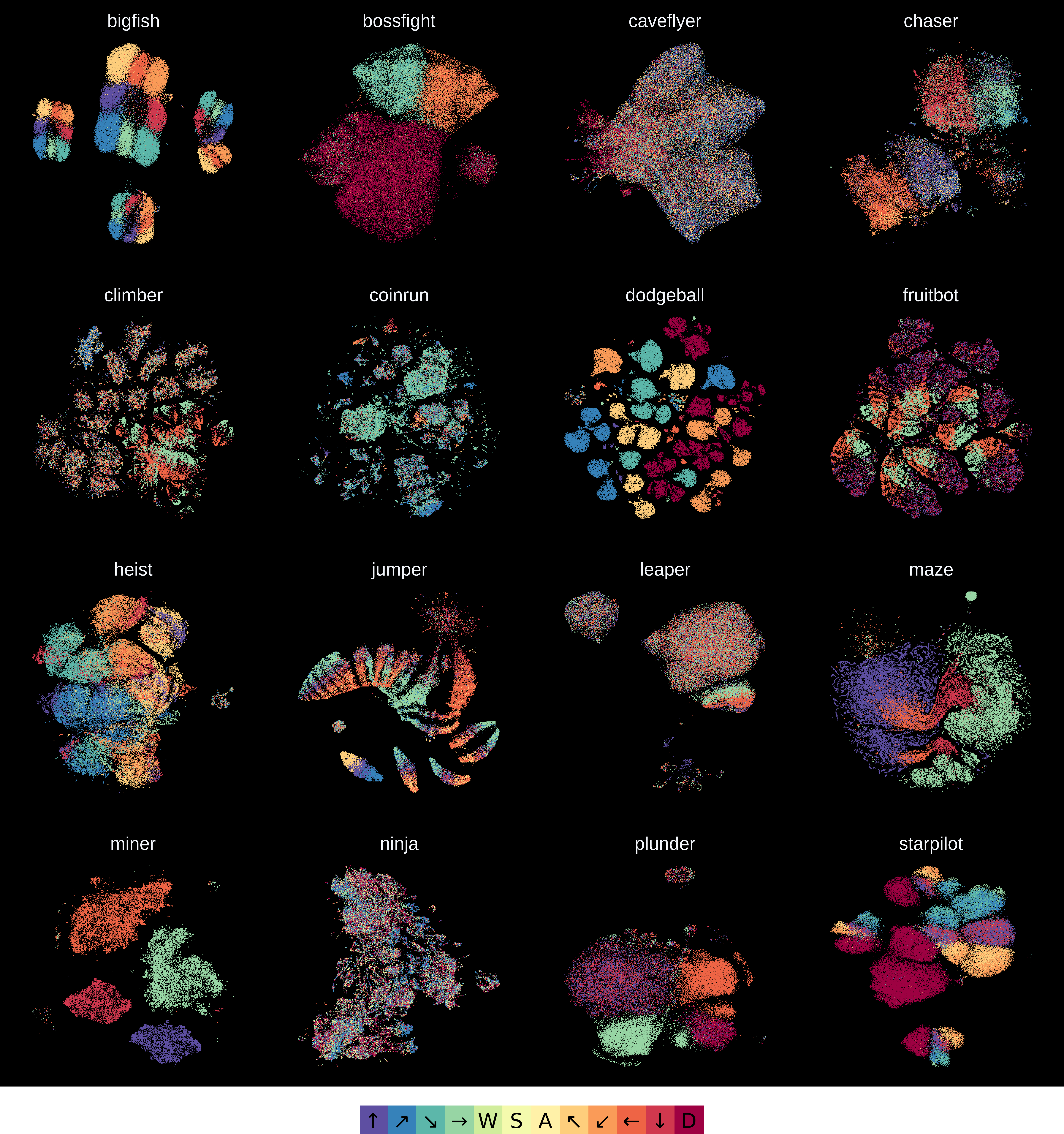}
\end{center}
\caption{UMAP projection of the learned latent action space for all 16 procgen games, generated by IDMs trained without vector-quantization. Each point represents the continuous latent action generated by the IDM for a transition in the observation-only dataset. Each point is color-coded by the true action taken by the agent at that transition (true action labels are only used for visualization, not for training). Arrows in the legend correspond to movement directions. \texttt{NOOP} actions are omitted for clarity.
\newline
We note that although the No-VQ ablation performs worse in terms of online performance, its FDM loss was significantly lower. This indicates that VQ indeed acts as a bottleneck that constrains the amount of information passed from the IDM to the FDM. By forcing the IDM to compress transition information, this leads to a better latent representation for downstream policy learning.
}
\end{figure}

\newpage
\subsection{Continuous control experiments}
\label{apdx:cc_results}

\begin{figure}[h]
\begin{center}
\includegraphics[width=1.0\textwidth]{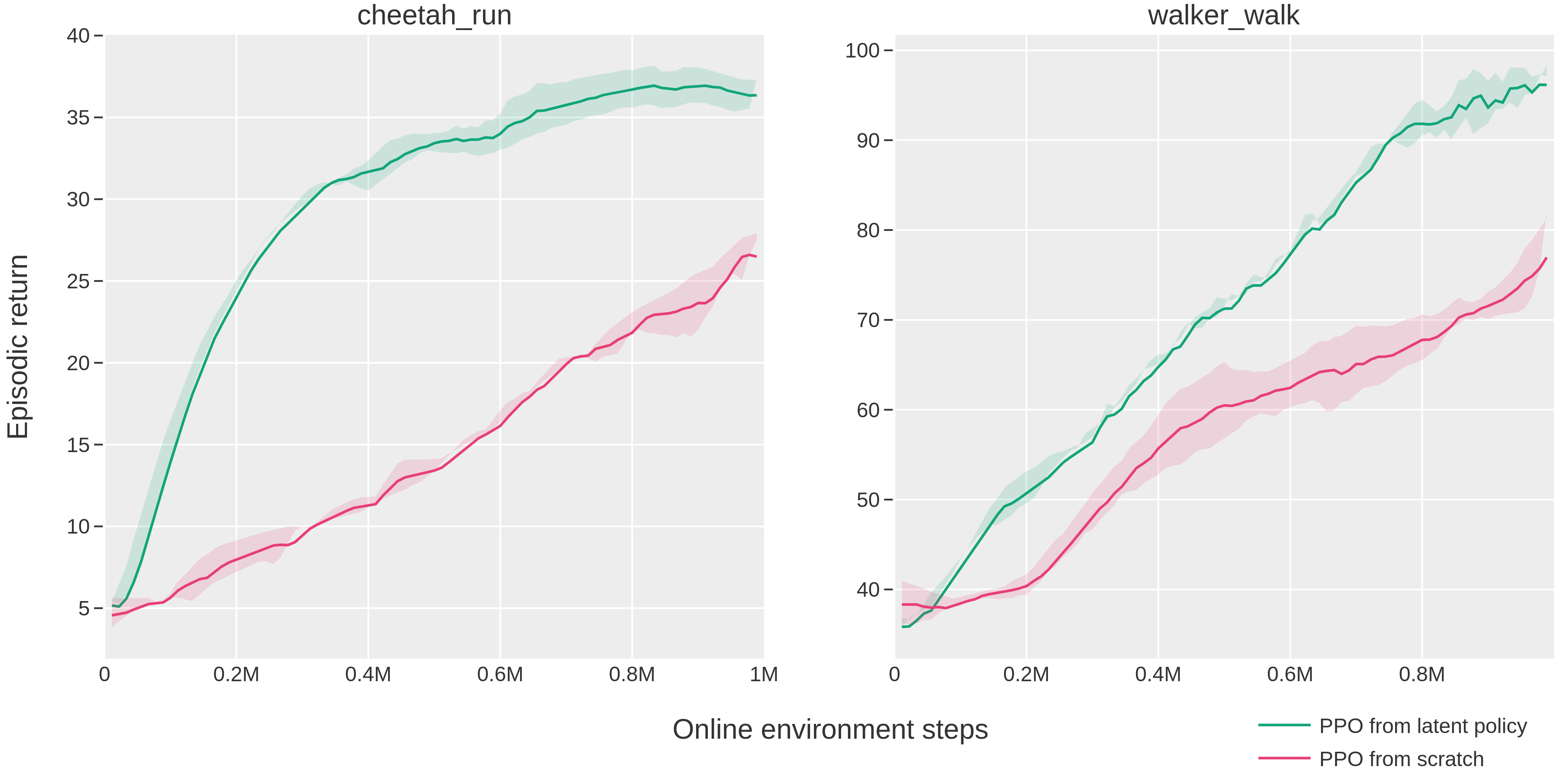}
\end{center}
\caption{Mean test returns over the course of training for online decoding of a \method{} latent policy compared to PPO from scratch (averaged across 2 seeds) on continuous control tasks. These experiments used the same hyperparameters for training the latent IDM and for behavior cloning as in Procgen experiments.}
\label{fig:offline_decoded_grid}
\end{figure}

\newpage
\subsection{Hyperparameters}
\label{apdx:hps}
\begin{table}[h]
\centering
\caption{Hyperparameters for the three stages of our method. Latent actions are 128-dimensional continuous vectors and are split and quantized into 8 discrete latents with 16-dimensional embeddings. We use Procgen distribution mode \texttt{easy}.}
\renewcommand{\arraystretch}{1.2} 
\begin{tabular}{lll}
Stage & Parameter & Value \\
\midrule
\multirow{11}{*}{Latent IDM training}       & Pre-transition additional context $k$ & 1 \\
                                            & VQ \# of codebooks & 2 \\
                                            & VQ \# of discrete latents per codebook & 4 \\
                                            & VQ \# of embeddings & 64 \\
                                            & VQ embedding dimension & 16 \\
                                            & VQ commitment cost & 0.05 \\
                                            & VQ EMA decay & 0.999 \\
                                            & Learning rate & 2e-4 \\
                                            & Batch size & 128 \\
                                            & Total update steps & 70,000 \\
                                            & IMPALA-CNN channel multiplier & 4 \\
\midrule
\multirow{4}{*}{Latent behavior cloning}    & Learning rate & 2e-4 \\
                                            & Batch size & 128 \\
                                            & Total update steps & 60,000 \\
                                            & IMPALA-CNN channel multiplier & 4 \\
\midrule
\multirow{2}{*}{RL online fine-tuning}      & Total environment interactions & 4,000,000 \\
                                            & PPO hyperparameters & As in \citep{procgen}. \\

\end{tabular}
\label{tab:hyperparameters}
\end{table}

\subsection{ILPO reproduction}
\label{apdx:ilpo_repro}

\begin{figure}[h]
\begin{center}
\includegraphics[width=0.6\textwidth]{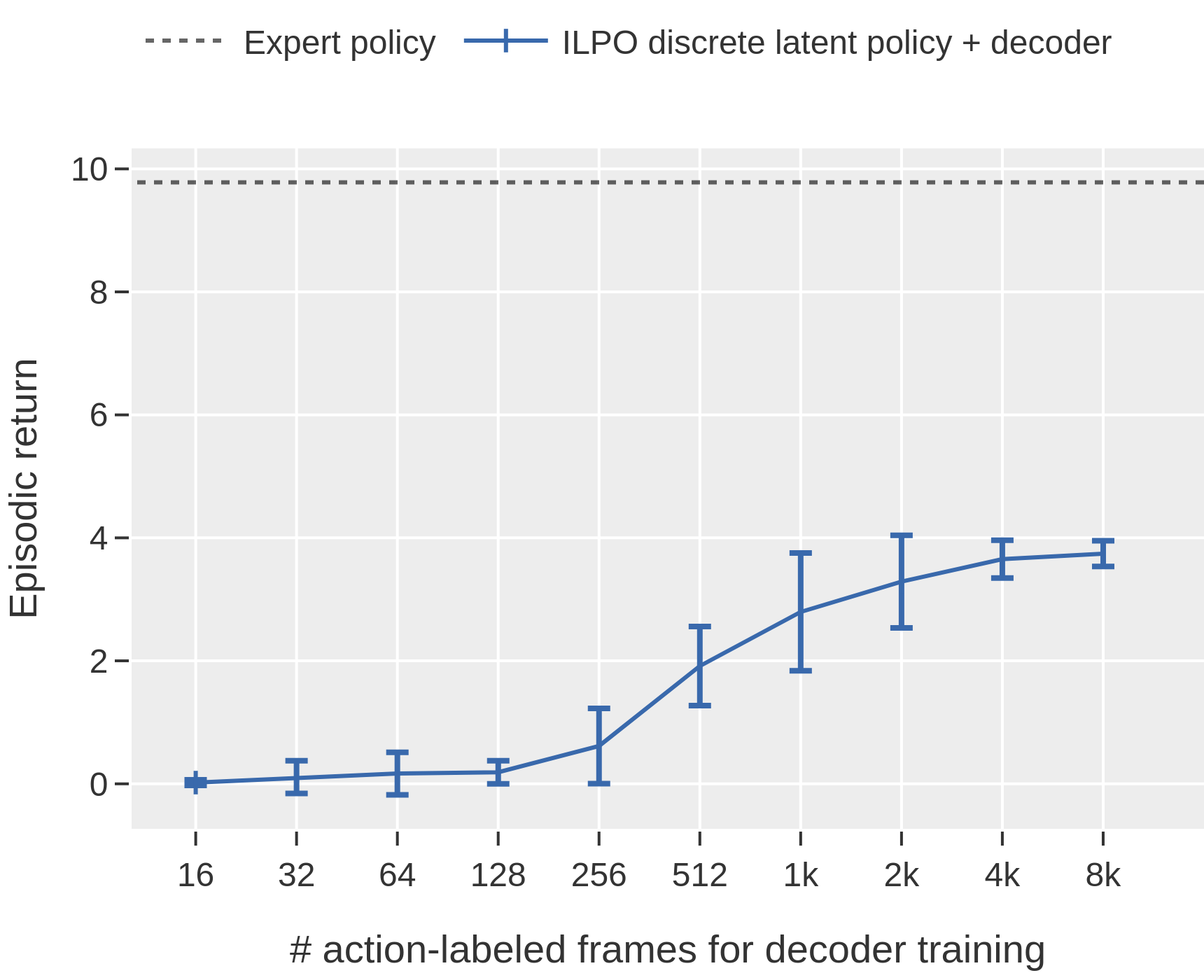}
\end{center}
\caption{Reproduction of Figure 5.a from \citet{ilpo} for a single level from \texttt{coinrun}. When applied to data from the full distribution of levels, rather than just a single level, the ILPO FDM consistently collapsed in terms of which discrete latent achieves the minimum of $\mathcal{L}_\text{min}$, leading to collapse of the policy too. Results here are not exactly comparable to results from \citet{ilpo} since it is unknown to us which specific \texttt{coinrun} level was used by the authors.}
\end{figure}

\end{document}